\documentclass[sigplan,screen]{acmart}
\renewcommand\footnotetextcopyrightpermission[1]{} 

\usepackage{enumitem} 

\usepackage{tikz}
\usetikzlibrary{arrows.meta,positioning}

\usepackage[T1]{fontenc}
\usepackage{url}

\AtBeginDocument{}

\setcopyright{acmlicensed}
\copyrightyear{2025}
\acmYear{2025}
\acmDOI{XXXXXXX.XXXXXXX}

\acmConference[ICAIF 2025 Workshop]{AI and Data Science for Digital Finance Workshop at ACM ICAIF 2025}{November 15, 2025}{Singapore}
\acmISBN{978-1-4503-XXXX-X/2025/11}

\title{ESGBench: A Benchmark for Explainable ESG Question Answering }

\author{Sherine George}
\affiliation{%
  \institution{BNY}
  \city{}
  \country{}
}
\email{sherinegeorge21@gmail.com}

\author{Nithish Saji}
\affiliation{%
  \institution{FedEx}
  \city{}
  \country{}
}
\email{nithish.nitt@gmail.com}

\begin{abstract}
Environmental, Social, and Governance (ESG) disclosures have become central to financial decision-making, yet sustainability reports remain lengthy, heterogeneous, and difficult to analyze at scale. Large Language Models (LLMs) show promise in extracting insights from these reports, but the field lacks standardized benchmarks to evaluate system accuracy, grounding, and explainability.

We introduce \textsc{ESGBench}\footnote{Code and dataset: \url{https://github.com/sherinegeorge21/ESGbench}}, an open benchmark for ESG-focused question answering (QA). ESGBench provides a reproducible pipeline to (i) collect ESG/TCFD reports, (ii) build a searchable index and table cache, (iii) automatically generate grounded QA pairs with verbatim evidence, and (iv) evaluate predictions across exact match, F1, numeric accuracy, and retrieval recall. Baseline experiments with a retrieval-augmented generation (RAG) system show modest exact match (21\%), string F1 (55\%), and numeric accuracy (45\%), underscoring the challenges of grounding numeric KPIs such as Scope~1--3 emissions, renewable energy, and diversity metrics. ESGBench enables systematic evaluation of ESG-specific QA systems and provides a foundation for research into explainable and trustworthy ESG analytics in digital finance.
\end{abstract}


\begin{document}
\maketitle

\section{Introduction}
Environmental, Social, and Governance (ESG) factors increasingly shape credit decisions, portfolio construction, and risk oversight. Yet the primary source of ESG information---corporate sustainability and TCFD reports---is long-form, heterogeneous PDF content that mixes narrative disclosures with numeric tables. This makes accurate, comparable extraction difficult for both analysts and machine systems. Large Language Models (LLMs) are promising but face two fundamental obstacles: (i) grounding and faithfulness when answering questions from long documents, and (ii) a lack of ESG-specific benchmarks to measure performance.

We introduce \textbf{ESGBench}, a small but reproducible benchmark for \emph{explainable} ESG question answering (QA). ESGBench provides (1) a public pipeline to ingest ESG/TCFD PDFs, (2) a chunked and table-aware index, (3) automatically generated QA pairs with verbatim evidence, and (4) an evaluation suite that reports exact match (EM), token F1, unit-aware numeric accuracy, retrieval recall@K, and per-category scores. A simple retrieval-augmented generation (RAG) baseline highlights current limitations: models struggle with numeric KPIs (e.g., Scope~1--3 emissions) and with table grounding.

Our goals are pragmatic: enable fast, comparable experiments; encourage evidence-grounded answers; and catalyze community extensions (more reports, human validation, multilingual coverage, and alignment to evolving standards such as CSRD and BRSR).

\section{Related Work}
\paragraph{Benchmarks for financial QA.}
Finance-oriented 
datasets (e.g., question answering over SEC filings and earnings calls) demonstrate LLM potential but focus on financial statements rather than ESG disclosures. ESGBench is complementary: it targets sustainability reports and emphasizes numeric environmental/social/governance KPIs with explicit evidence.

\paragraph{Domain models and climate text.}
Specialized encoders such as FinBERT and ClimateBERT improve representation learning for financial or climate corpora. ESGBench is model-agnostic: it evaluates any system (LLMs, RAG, rule-based extractors) using uniform metrics and transparent evidence.

\paragraph{RAG for factuality and explainability.}
Retrieval-
augmented generation improves grounding by constraining answers to retrieved context. Prior work rarely reports \emph{retrieval recall} against human-labeled evidence. ESGBench makes recall@K first-class and requires verbatim evidence spans for each gold answer.

\section{Dataset and Pipeline}
\subsection{Document Collection}
We seed the corpus with widely-available reports from large, publicly-known companies across sectors (technology, consumer, energy, finance, manufacturing). Each entry in \\ \texttt{data/docs\_seed.csv} records \textit{company, year, url, doc\_type, country, industry, source}. The ingestion script then downloads the PDFs to \texttt{pdfs/} and logs status and 
checksums in \\ \texttt{data/esgbench\_document\_information.jsonl} respectively. 
\\ Dead/forbidden links are retained as metadata but skipped downstream.

\subsection{Preprocessing}
We create two views:
\begin{itemize}[leftmargin=*, itemsep=1pt, topsep=3pt]
  \item \textbf{Passage chunks} (\texttt{cache/chunks.json}): PDFs are text-extracted and chunked (600–1200 chars) with page numbers.
  \item \textbf{Table rows} (\texttt{cache/\{DOC\}\_tables.json}): Tables parsed via pdfplumber/Camelot; rows normalized with optional \texttt{value}/\texttt{unit}.
\end{itemize}

Both views retain document names and page indices so that QA evidence is auditable.

\subsection{QA Generation}
We use a constrained prompting scheme to produce QA pairs from either (i) passage chunks (narrative facts) or (ii) compact table-row summaries (numeric KPIs). Prompts enforce: (a) questions must be answerable \emph{only} from the provided text, (b) answers should be \emph{verbatim} (numbers and units preserved), and (c) an \texttt{evidence\_quote} must be copied verbatim. We de-duplicate by hashing \texttt{(doc\_name | question | answer)}. The final gold file \texttt{data/esgbench\_open\_source.json} stores company, doc, category (Environmental\slash Social\slash Governance\slash Strategy\slash Risk), KPI name, question, answer, and evidence with page number.
\subsection{Dataset Statistics}
Table~\ref{tab:stats} summarizes the initial release. The mix intentionally includes multiple regions and report formats to stress robustness. ESGBench is small by design but easily extensible via the public scripts.

\begin{table}[t]
\centering
\caption{ESGBench v0.1: corpus profile. Replace counts with your latest run.}
\label{tab:stats}
\begin{tabular}{@{}lr@{}}
\toprule
Companies & 10 \\
Reports (PDFs) & 12 \\
Total QA pairs & 119 \\
Avg. QAs / report & 9.9 \\
Category split (E / S / G / Strategy / Risk) & 50 / 14 / 23 / 11 / 2 \\
Table-derived QAs (\%) & 40--50\% \\
\bottomrule
\end{tabular}
\end{table}

\section{Baseline Model}
We implement a simple RAG system to ground answers:
\begin{itemize}[leftmargin=*how to, itemsep=1pt, topsep=3pt]
  \item \textbf{Retriever:} Dense embeddings over chunks using a general-purpose embedding model; top-$k$ retrieval per question.
  \item \textbf{Reader:} A constrained generation prompt asks for a \emph{single} span answer drawn from the retrieved text; for numeric questions we instruct the model to preserve digits and units verbatim.
  \item \textbf{Prediction schema:} Each record contains \\ \texttt{doc\_name}, \texttt{question}, \texttt{pred}, and \texttt{retrieved\_pages} for \\recall@K.
\end{itemize}

\section{Evaluation}
We report four complementary metrics:
\begin{itemize}[leftmargin=*, itemsep=1pt, topsep=3pt]
  \item \textbf{Exact Match (EM):} case-insensitive string equality.
  \item \textbf{String F1:} token F1 after light normalization.
  \item \textbf{Numeric Accuracy@$\pm$2\%:} unit-aware tolerance; numbers must match within 2\% and units (if present) must align.
  \item \textbf{Retrieval Recall@K:} whether a gold evidence page is among retrieved pages.
\end{itemize}

Per-category accuracy (Environmental, Social, Governance, etc.) helps identify systematic weaknesses.

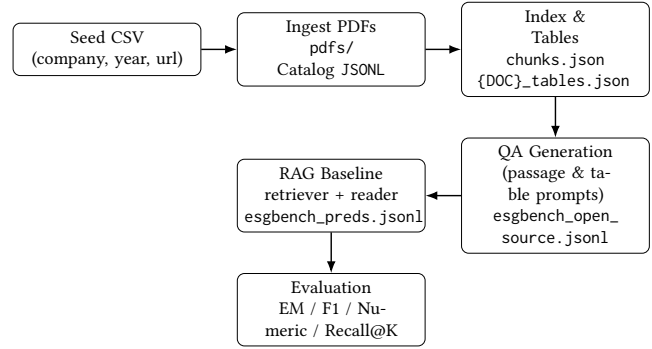
\begin{figure}[t]
  \centering
  \resizebox{\linewidth}{!}{%
  \begin{tikzpicture}[
      node distance = {8mm and 7mm},
      proc/.style = {rectangle, rounded corners, draw, align=center,
                     inner sep=4pt, text width=3.4cm}, 
      arrow/.style = {-Latex, thick}
  ]
    \node[proc] (seed)   {Seed CSV\\(company, year, url)};
    \node[proc, right=of seed] (ingest) {Ingest PDFs\\\texttt{pdfs/}\\Catalog \texttt{JSONL}};
    \node[proc, right=of ingest] (index) {Index \&\\Tables\\\texttt{chunks.json}\\\texttt{\{DOC\}\_tables.json}};
    \node[proc, below=of index] (qa) {QA Generation\\(passage \& table prompts)\\\texttt{esgbench\_open\_
    source.jsonl}};
    \node[proc, left=of qa]  (rag) {RAG Baseline\\retriever + reader\\\texttt{esgbench\_preds.jsonl}};
    \node[proc, below=of rag] (eval) {Evaluation\\EM / F1 / Numeric / Recall@K};

    \draw[arrow] (seed)  -- (ingest);
    \draw[arrow] (ingest)-- (index);
    \draw[arrow] (index) -- (qa);
    \draw[arrow] (qa)    -- (rag);
    \draw[arrow] (rag)   -- (eval);
  \end{tikzpicture}%
  }
  \caption{ESGBench pipeline}
  \label{fig:pipeline}
\end{figure}

\section{Results}
Table~\ref{tab:baseline} reports baseline performance averaged over the current gold set. We observe: (i) governance and strategy questions are comparatively easier (short factual spans), (ii) environmental numerics from tables are error-prone (formatting, unit conversion, scale words like ``thousand''), and (iii) recall gaps propagate to answer errors, supporting the need to report recall@K alongside generation metrics.

\begin{table}[t]
\centering
\caption{RAG baseline on ESGBench v0.1 }
\label{tab:baseline}
\begin{tabular}{l r}
\toprule
Evaluated QA pairs & 119 \\
Exact Match (EM) & 21.0\% \\
String F1 (avg) & 55.4\% \\
Numeric Accuracy@$\pm$2\% & 45.3\% \\
Retrieval Recall@5 (if logged) & 70--80\% \\
\midrule
Environmental (EM/Numeric OK) & 48.0\% \\
Governance & 43.5\% \\
Strategy & 90.9\% \\
Social & 35.7\% \\
\bottomrule
\end{tabular}
\end{table}

\section{Discussion and Limitations}
\paragraph{Coverage.} The seed corpus favors large multinationals with English-language reports; this biases topics and style.  

\paragraph{Generation noise.} Although prompts enforce verbatim evidence, automatic QA generation can miss context or prefer easily extractable spans. ESGBench mitigates this by requiring verbatim evidence strings and by reporting recall@K.  

\paragraph{Tables and units.} Variants like ``tCO\textsubscript{2}e,'' ``ktCO\textsubscript{2}e,'' and scale words (``million'') create fragile failure modes.  

\paragraph{Ethics and governance.} ESGBench does not score ``truthfulness'' of corporate claims—only the system’s ability to recover statements from disclosures.

\paragraph{Future work.}
(i) Human validation of a stratified QA subset, (ii) multilingual reports, (iii) alignment to CSRD/BRSR/ISSB taxonomies for coverage metrics, (iv) layout-aware table parsing, and (v) robustness tests for prompt sensitivity and retrieval noise.

\section{Conclusion}
ESGBench is a compact, reproducible benchmark for explainable ESG QA. By releasing ingestion, indexing, QA generation, RAG baseline, and evaluation code, we lower the barrier to rigorous, evidence-grounded experiments on sustainability disclosures. We invite contributions expanding document coverage, adding curated human-validated QAs, and exploring methods that improve numeric grounding and table reasoning.

\begin{acks}
We thank contributors to the ESGBench repository and colleagues for helpful discussions.
\end{acks}

\bibliographystyle{ACM-Reference-Format}

\end{document}